\documentclass{article}
\usepackage{float}
\usepackage{tabularx}
\usepackage{PRIMEarxiv}
\usepackage{natbib}
\usepackage[utf8]{inputenc} 
\usepackage[T1]{fontenc}    
\usepackage{hyperref}       
\usepackage{url}            
\usepackage{booktabs}       
\usepackage{amsfonts}       
\usepackage{nicefrac}       
\usepackage{microtype}      
\usepackage{lipsum}
\usepackage{fancyhdr}       
\usepackage{graphicx}       
\graphicspath{{media/}}     
\usepackage{longtable}
\usepackage{float}
\usepackage{enumitem}
\usepackage{amsmath}
\usepackage{graphicx}
\usepackage{rotating}
\usepackage{array}
\usepackage{makecell}

\title{GLiClass: Generalist Lightweight Model for Sequence Classification Tasks
}
\author{
\textbf{Ihor Stepanov$^{1}$, Mykhailo Shtopko$^{1}$, Dmytro Vodianytskyi$^{1}$,}\\
\textbf{Oleksandr Lukashov$^{1}$, Alexander Yavorskyi$^{1}$, Mykyta Yaroshenko$^{1}$}\\
$^{1}$Knowledgator Engineering, Kyiv, Ukraine \\
\textbf{Correspondence:} \texttt{ingvarstep@knowledgator.com}, \texttt{mykhailoshtopko@knowledgator.com}
}

\begin{document}
\maketitle
\begin{abstract}
Classification is one of the most widespread tasks in AI applications, serving often as the first step in filtering, sorting, and categorizing data. Since modern AI systems must handle large volumes of input data and early pipeline stages can propagate errors downstream, achieving high efficiency and accuracy is critical. Moreover, classification requirements can change dynamically based on user needs, necessitating models with strong zero-shot capabilities. While generative LLMs have become mainstream for zero-shot classification due to their versatility, they suffer from inconsistent instruction following and computational inefficiency. Cross-encoders, commonly used as rerankers in RAG pipelines, face a different bottleneck: they must process text-label pairs sequentially, significantly reducing efficiency with large label sets. Embedding-based approaches offer good efficiency but struggle with complex scenarios involving logical and semantic constraints. We propose GLiClass, a novel method that adapts the GLiNER architecture for sequence classification tasks. Our approach achieves strong accuracy and efficiency comparable to embedding-based methods, while maintaining the flexibility needed for zero-shot and few-shot learning scenarios. Additionally, we adapted proximal policy optimization (PPO) for multi-label text classification, enabling training classifiers in data-sparse conditions or from human feedback.
\end{abstract}


\section{Introduction}\label{sec:introduction}

Text classification is a fundamental task in machine learning with an extensive research history and significant practical applications \citep{li2022survey}. It serves as a critical component in information extraction and analytical systems, powering diverse applications from scientific article categorization \citep{lee2020biobert} and support ticket classification \citep{revina2020ticket} to sentiment analysis \citep{giachanou2016like} and financial research \citep{Felgueiras2020CreatingCM}. When generalized to sequence classification, the impact extends further, including DNA sequence analysis \citep{helaly2022bert} and RAG pipelines \citep{Rosa2022InDO}, which have become essential for ensuring up-to-date, high-quality outputs in modern chatbot systems.

Recent advances in auto-regressive language modeling have opened new possibilities for zero-shot classification tasks \citep{brown2020languagemodelsfewshotlearners, raffel2023exploringlimitstransferlearning}, including text classification \citep{puri2019zero, rasheed2024codepori}. Although these models demonstrate impressive versatility, they often struggle with strict instruction adherence and suffer from computational inefficiency in training and inference phases.

Cross-encoders operating as Natural Language Inference (NLI) models represent another popular approach for zero-shot classification \citep{yin2019benchmarkingzeroshottextclassification, laurer2023building} and RAG pipelines \citep{Rosa2022InDO}. These models treat the input sequence as an NLI premise and construct hypotheses from candidate labels. Although more computationally efficient than LLMs, they face scalability challenges with large label sets due to their pairwise processing requirement. Furthermore, their limited ability to comprehend cross-label information can affect the quality of prediction in complex scenarios.

Since the introduction of Word2Vec \citep{mikolov2013efficientestimationwordrepresentations}, embedding-based approaches have emerged as efficient methods for text classification \citep{su2014chinese}, particularly in zero-shot settings \citep{Dai2019MultilingualWE}. The development of sentence encoders improved semantic understanding \citep{perone2018evaluationsentenceembeddingsdownstream}, and Sentence Transformers \citep{reimers-2019-sentence-bert} further enhanced embedding quality, enabling classification without fine-tuning \citep{Piao2021ScholarlyTC}. SetFit extended this approach to achieve strong performance with minimal training examples \citep{tunstall2022efficientfewshotlearningprompts}. Despite their efficiency, embedding-based methods often fall short in complex scenarios involving logical and semantic constraints.

This work introduces GLiClass, a novel sequence classification model inspired by the GLiNER architecture \citep{zaratiana2023glinergeneralistmodelnamed} and explicitly adapted for text classification tasks. While developing the first multi-task GLiNER model \citep{stepanov2024gliner}, text classification emerged as one of the evaluated tasks, exposing limitations that highlighted the need for a more specialized solution. GLiClass addresses these limitations by combining the accuracy of advanced architectures with the efficiency of embedding-based methods, while preserving strong zero-shot and few-shot generalization capabilities. In this paper, we present updated architectural variants of GLiClass along with an enhanced training methodology. The resulting models achieve performance comparable to or exceeding that of cross-encoder baselines, with significantly improved computational efficiency. The development of other GLiNER derivatives—such as GLiREL \citep{boylan2025glirel} for zero-shot relation extraction and GLiDRE \citep{armingaud2025glidre} for document-level relation extraction further demonstrates the flexibility of the GLiNER framework, motivating the creation of task-specific generalist models like GLiClass.
\section{Methods}\label{seq:methods}

\subsection{Model Architecture}
We developed several variants of the GLiClass architecture, with the main version built upon the GLiNER uni-encoder design. Our primary models use the DeBERTa backbone \citep{He2020DeBERTaDB}, specifically DeBERTa v3 \citep{He2021DeBERTaV3ID}, which incorporates Electra-style pretraining \citep{clark2020electra}—an approach shown to be particularly effective for text classification tasks. In addition, we experimented with models based on the ModernBERT backbone \citep{modernbert, weller2025seqvsseqopen}, which integrates several modern architectural enhancements, including support for Flash Attention \citep{Dao2022FlashAttentionFA} and an extended context window. Despite these advancements, our experiments indicate that DeBERTa-based models consistently outperform those based on ModernBERT. The GLiClass architecture was designed to meet the following objectives:
\begin{itemize}
\item Perform multi-label classification in a single forward-pass, enabling efficient handling of multiple categories without repeated computations;
\item Achieve non-linear scaling with the number of classes provided, ensuring that inference time does not increase proportionally with label count, which is crucial for large-scale applications;
\item Enable inter-label information communication, allowing the model to capture relationships, hierarchies, and dependencies between labels to improve prediction quality in complex scenarios.
\end{itemize}
At the time, while making GLiCLass more computationally efficient, our goal was to achieve performance at the level of cross-encoders or even better accuracy, especially in the cases where inter-label communication can help.

To achieve this, a uni-encoder architecture was selected as the primary design, where text and labels are processed jointly in a single encoder to facilitate rich interactions; however, we also developed and explored other variants, such as bi-encoder (separate encoders for text and labels), fused bi-encoder (combining embeddings early), and encoder-decoder (with cross-attention mechanisms), each offering trade-offs in efficiency, flexibility, and performance for different use cases.

\subsubsection{Architecture Overview}

\begin{figure}[H]
    \centering
    \includegraphics[width=1\linewidth]{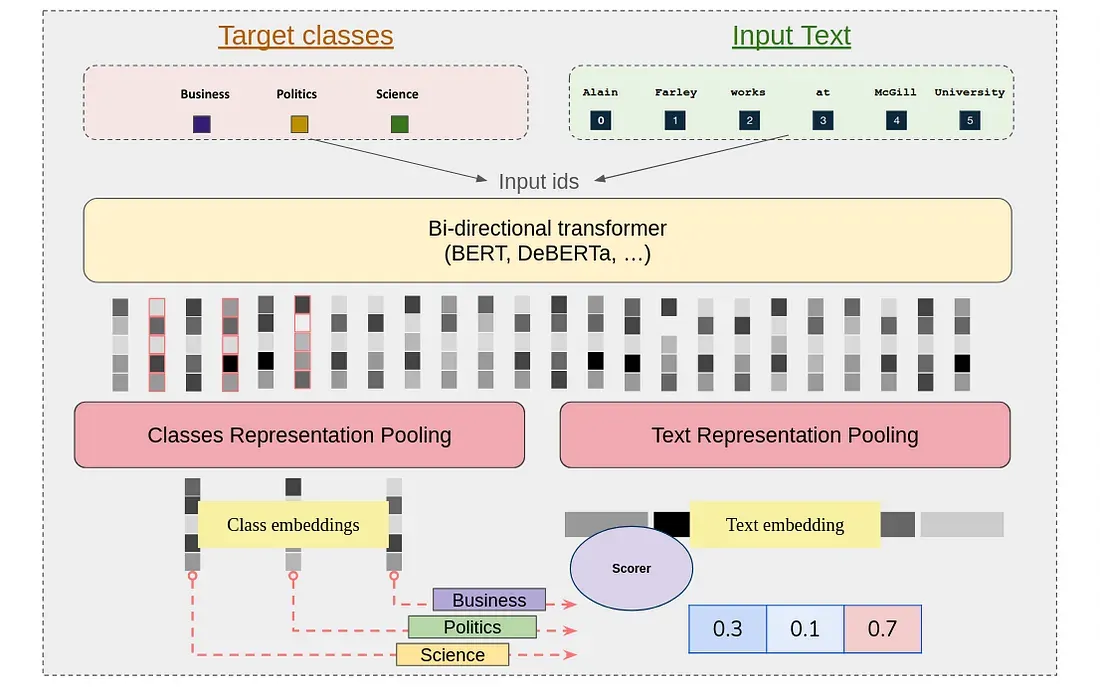}
    \caption{GLiClass Uni-Encoder Architecture}
    \label{fig:gliclass-uni-encoder-scheme}
\end{figure}

GLiClass employs a sequence classification architecture that jointly processes label tokens with input text to enable rich text-label interactions while maintaining computational efficiency. The pipeline consists of four main stages: (i) input and label integration, (ii) contextual representation learning, (iii) representation pooling, and (iv) scoring.

\subsubsection{Input Processing and Label Integration}

Each class label is prepended with a special token \texttt{<<LABEL>>} and concatenated with the input text. This construction allows the encoder to process text and labels jointly while preserving their distinct semantic roles.

\subsubsection{Contextual Representation Learning}

The concatenated sequence is processed through a bidirectional transformer encoder (e.g., BERT or DeBERTa), which facilitates:
\begin{itemize}
    \item Label-label interactions (capturing relations and hierarchies)
    \item Text-label interactions (text informing label representations)
    \item Label-text interactions (labels guiding text interpretation)
\end{itemize}

Unlike pairwise cross-encoders, this joint processing captures inter-label dependencies that would otherwise be missed, leading to more informed predictions.

\subsubsection{Representation Pooling}

From the encoder outputs, we extract text and label representations separately using one of three pooling strategies:
\begin{itemize}
    \item First-token pooling
    \item Mean pooling
    \item Attention-weighted pooling
\end{itemize}
The pooling strategy can be selected based on task requirements.

\subsubsection{Scoring Mechanism}

Let $\mathbf{t}\in\mathbb{R}^{B\times D}$ denote the pooled text embedding and $\mathbf{c}\in\mathbb{R}^{B\times C\times D}$ the pooled label embeddings for $C$ classes. We compute logits using either dot product or a learnable scorer:

\begin{align}
\text{Dot Product:} \quad s_{b,k} &= \frac{\mathbf{t}_b^\top \mathbf{c}_{b,k}}{\tau} \\
\text{NN Scorer:} \quad s_{b,k} &= g([\mathbf{t}_b;\mathbf{c}_{b,k}])
\end{align}

where $\tau > 0$ is a temperature parameter and $g(\cdot)$ is a small MLP.

\subsubsection{Layer-wise Attention Re-weighting}

To optimize information flow across encoder layers, we implement a squeeze-excitation scheme. Let encoder layer outputs be $\{U^{(k)}\}_{k=1}^{K}$, with $U^{(k)}\in\mathbb{R}^{B\times L\times D}$. We compute layer weights as:

\begin{align}
Z_{:,k} &= \frac{1}{L}\sum_{l=1}^{L}\text{Linear}_{\mathrm{squeeze}}\left(U^{(k)}_{:,l,:}\right) \in \mathbb{R}^{B\times K} \\
S &= \sigma\left(W_2\,\mathrm{ReLU}(W_1 Z^\top)\right)^\top \in \mathbb{R}^{B\times K} \\
\tilde{U} &= \sum_{k=1}^{K} S_{:,k} \cdot U^{(k)} \in \mathbb{R}^{B\times L\times D} \\
O &= \text{Linear}_{\mathrm{proj}}(\tilde{U}) \in \mathbb{R}^{B\times L\times D_{\mathrm{out}}}
\end{align}

where $W_1 \in \mathbb{R}^{\frac{K}{2}\times K}$ and $W_2 \in \mathbb{R}^{K\times \frac{K}{2}}$.

\subsubsection{Token-level Contrastive Loss}

To enhance representation quality, we employ a token-level contrastive loss. Given embeddings $E\in\mathbb{R}^{B\times L\times D}$ and token mask $M\in\{0,1\}^{B\times L}$, let $\hat{E}=E/\|E\|_2$ be the L2-normalized embeddings along $D$. For each batch $b$, the similarity matrix is:

\begin{equation}
S^{(b)} = \hat{E}_b\,\hat{E}_b^\top \in \mathbb{R}^{L\times L}
\end{equation}

The contrastive loss trains each valid token to identify itself among all tokens in its sequence:

\begin{equation}
\mathcal{L} = \frac{1}{\sum_{b,l} M_{b,l}}\sum_{b=1}^{B}\sum_{l=1}^{L} M_{b,l} \cdot \mathrm{CE}\big(S^{(b)}_{l,:}, l\big)
\end{equation}

\subsubsection{Architectural Variants}

We explore four architectural configurations, each with specific advantages:

\paragraph{Uni-encoder:} Processes text and labels jointly through a single encoder:
\begin{align}
\mathbf{H} &= \text{Encoder}(\mathbf{X}) \\
\mathbf{C}, \mathbf{M}_c &= \text{ExtractClassFeatures}(\mathbf{H}, \text{class\_tokens}) \\
\mathbf{T} &= \text{Pooler}(\mathbf{H}) \\
\text{Logits} &= \text{Scorer}(\mathbf{T}, \mathbf{C})
\end{align}

\paragraph{Bi-encoder:} Uses separate encoders for text and labels:
\begin{align}
\mathbf{T} &= \text{TextEncoder}(\mathbf{X}_{\mathrm{text}}) \\
\mathbf{C} &= \text{ClassEncoder}(\mathbf{X}_{\mathrm{class}}) \\
\text{Logits} &= \text{Scorer}(\mathbf{T}, \mathbf{C})/\tau
\end{align}

\paragraph{Fused bi-encoder:} Combines class embeddings with text at the embedding layer:
\begin{align}
\mathbf{C}_{\mathrm{raw}} &= \text{ClassEncoder}(\mathbf{X}_{\mathrm{class}}) \\
\mathbf{E} &= \text{EmbeddingLayer}(\mathbf{X}_{\mathrm{text}}) \\
\mathbf{E}[\text{class\_token\_pos}] &= \mathbf{C}_{\mathrm{raw}}[\text{selected\_classes}] \\
\mathbf{H}, \mathbf{C}_{\mathrm{fused}} &= \text{TextEncoder}(\mathbf{E}) \\
\text{Logits} &= \text{Scorer}(\text{Pool}(\mathbf{H}), \mathbf{C}_{\mathrm{fused}})
\end{align}

\paragraph{Encoder-decoder:} Employs an encoder-decoder architecture with cross-attention:
\begin{align}
\mathbf{H}_{\mathrm{enc}} &= \text{Encoder}(\mathbf{X}_{\mathrm{text}}) \\
\mathbf{H}_{\mathrm{dec}} &= \text{Decoder}(\mathbf{X}_{\mathrm{class}}, \mathbf{H}_{\mathrm{enc}}) \\
\mathbf{C} &= \text{ExtractClassFeatures}(\mathbf{H}_{\mathrm{dec}}) \\
\text{Logits} &= \text{Scorer}(\text{Pool}(\mathbf{H}_{\mathrm{enc}}), \mathbf{C})
\end{align}

\subsection{Data}\label{sec:data}

\paragraph{Pre-training corpus:}
A 1.2M example general-purpose dataset covering text classification, sentiment analysis, and natural language inference tasks.

\paragraph{Mid-training corpus:}
A representative subset of the pre-training corpus, used for intermediate fine-tuning.

\paragraph{Logic/NLI stream (post-training):}
Logical reasoning datasets including \texttt{tau/CommonsenseQA} and 2,000 synthetic examples covering formal logic, sequent calculus, and NLI-style entailment/contradiction.

\paragraph{Pattern-focused stream (post-training):}
To address length and label density patterns, we created a dataset with texts grouped by word count:
$[0, 4, 8, 16, 24, 32, 48, 64, 96, 128, 192, 256, 384, 512, 768, 1024]$

Short buckets (0-8 words) were populated with short \texttt{["title"]} fields; buckets 8-48 with \texttt{fancyzhx/amazon\_polarity} using the \texttt{["content"]} field; and buckets 48-1024 with samples from \texttt{m-a-p/FineFineWeb}. All buckets were filled equally. Each text was annotated with GPT-4o to generate 50 true and 50 false candidate labels. For the final pattern-focused set, we sampled 2,000 texts in equal proportions from all buckets; each example was duplicated, and for the duplicate, we varied the number of positive/negative labels using random coefficients to diversify label density.

\paragraph{Additional NLI:}
Examples from \texttt{nyu-mll/MultiNLI} to strengthen classical NLI capabilities.

\subsection{Model Training}

\subsubsection{Training Framework}

\textbf{Data Preparation:} The dataset is loaded from JSON format, randomly shuffled, and split into 90\% training and 10\% test partitions. Input sequences are tokenized with a maximum length of 1024 tokens using dynamic padding.

We implement two complementary training pipelines: standard supervised learning using focal loss and reinforcement learning (RL), both extending the Hugging Face \texttt{Trainer} framework. The RL pipeline employs a modified Proximal Policy Optimization (PPO) approach adapted for text classification.

\subsubsection{Reinforcement Learning Loss Function}

The total loss combines four components:

\begin{equation}
\mathcal{L}_{\text{total}} = \mathcal{L}_{\text{PPO}} + \mathcal{L}_{\text{value}} + \mathcal{L}_{\text{KL}} + \mathcal{L}_{\text{entropy}}
\end{equation}

\paragraph{1. PPO Loss:}
\begin{equation}
\mathcal{L}_{\text{PPO}} = -\frac{1}{N}\sum_{i=1}^N\sum_{j=1}^L \min\Big( r_{ij}\hat{A}_{ij}, \text{clip}(r_{ij}, 1\pm\epsilon)\hat{A}_{ij}\Big)
\end{equation}

where $r_{ij} = \frac{\pi_\theta(a_{ij}|s_i)}{\pi_{\theta_{\text{old}}}(a_{ij}|s_i)}$ is the probability ratio between current and previous policies, and $\hat{A}_{ij} = R_{ij} - V(s_i)$ is the advantage estimate.

\paragraph{2. Value Loss:}
\begin{equation}
\mathcal{L}_{\text{value}} = |V(s) - R|^2
\end{equation}
This measures the accuracy of the value model's reward prediction.

\paragraph{3. KL-Divergence Penalty:}
\begin{equation}
\mathcal{L}_{\text{KL}} = \beta_{\text{KL}} D_{\text{KL}}(\pi_{\text{ref}} \parallel \pi_\theta)
\end{equation}
Penalizes deviation from a reference policy, controlled by coefficient $\beta_{\text{KL}}$.

\paragraph{4. Entropy Bonus:}
\begin{equation}
\mathcal{L}_{\text{entropy}} = \beta_{\text{ent}} \mathbb{H}(\pi_\theta)
\end{equation}
Encourages policy exploration through prediction uncertainty, weighted by $\beta_{\text{ent}}$.

\paragraph{Key Hyperparameters:}
\begin{itemize}
    \item \textit{Clip range} ($\epsilon$): 0.2 (constrains policy updates)
    \item \textit{RL iterations}: 3 (updates per batch)
    \item \textit{Entropy coefficient} ($\beta_{\text{ent}}$): -1 (disabled by default)
    \item \textit{KL coefficient} ($\beta_{\text{KL}}$): -1 (disabled by default)
    \item \textit{Focal loss $\alpha$}: -1 (disabled)
    \item \textit{Focal loss $\gamma$}: -1 (disabled)
    \item \textit{Label smoothing}: -1 (disabled)
\end{itemize}

\paragraph{Special Modifications:}
\begin{itemize}
    \item \textit{Focal Loss Adaptation:}
    \begin{equation}
    \mathcal{L}_{\text{PPO}} \leftarrow \frac{1}{N}\sum_{i,j} \alpha_{ij} (p_{ij}^t)^\gamma \mathcal{L}_{\text{PPO}}
    \end{equation}
    
    \item \textit{Label Smoothing:}
    \begin{equation}
    a_{ij}^{\text{smooth}} = (1 - \epsilon_{\text{smooth}})a_{ij} + 0.5\epsilon_{\text{smooth}}
    \end{equation}
    
    \item \textit{Action Sampling:}
    \begin{equation}
    a_{ij} \sim \begin{cases}
    \text{Bernoulli}(p_{ij}) & \textit{(stochastic)} \\
    \mathbb{I}(p_{ij} > 0.5) & \textit{(deterministic)}
    \end{cases}
    \end{equation}
\end{itemize}

\paragraph{Training Execution:}
\begin{enumerate}
    \item \textit{Multi-iteration Updates}: Each batch undergoes $N_{\text{RL-iters}}$ policy refinements
    \item \textit{Separate Optimizers}: Policy ($\pi_\theta$) and value ($V_\phi$) models have dedicated AdamW optimizers
    \item \textit{Reference Integration}: Frozen zero-shot pipeline provides baseline probabilities $\pi_{\text{ref}}$
    \item \textit{Reward Composition}: $R = \sum_i w_i r_i(s,a)$ with configurable components
    \item \textit{Monitoring}: Tracks $\mathcal{L}_{\text{total}}$, $\mathbb{E}[\hat{A}]$, and individual reward metrics
\end{enumerate}

\paragraph{Shared Infrastructure:}
\begin{itemize}
    \item \textit{Layer-specific Optimization}: Encoder layers use $\eta = 10^{-5}$, $\delta = 0.01$; classifier layers use $\eta_{\text{other}} = 3\times10^{-5}$, $\delta_{\text{other}} = 0.01$
    \item \textit{Gradient Handling}: LayerNorm parameters excluded from weight decay
    \item \textit{Fault Tolerance}: Per-batch exception handling with cache clearing
    \item \textit{Checkpointing}: Snapshots every 1000 steps with 3-checkpoint retention
\end{itemize}

\subsubsection{Training Stages}

\paragraph{Pre-Training:}
Initial training on the 1.2M example corpus to learn general classification patterns and train custom class tokens for pooling representations. Post hoc inspection of attention scores revealed two issues: (i) as the number of labels increases, attention between tokens of labels and label tokens (prefixed with \texttt{<<LABEL>>}) diminishes; (ii) under extreme label-to-text token ratios (many labels and short texts), text representations degrade.

\paragraph{Mid-Training:}
Intermediate fine-tuning using the RL trainer on a subset of the pre-training corpus to refine decision boundaries and improve label-text alignment. This bridge between large-scale pre-training and targeted post-training yielded modest but consistent gains in macro F1 across diverse datasets.

\paragraph{Post-Training:}
Final stage combining logic/NLI and pattern-focused streams using Low-Rank Adaptation (LoRA) to preserve prior knowledge while adapting to new patterns. We found that fine-tuning GLiClass on formal-logic tasks formulated as question answering and classical NLI improves zero-shot text classification. The \texttt{edge} variant trained more stably when using higher-rank (over-parameterized) LoRA adapters. Table~\ref{tab:lora-all-transposed} shows the LoRA configurations for each model variant.

\begin{table}[ht]
\centering
\small
\caption{LoRA configuration for GLiClass post-training}
\setlength{\tabcolsep}{6pt}
\begin{tabular}{lccccl}
\toprule
Model & LoRA rank $r$ & LoRA $\alpha$ & Focal loss $\alpha$ & Target modules \\
\midrule
\texttt{gliclass-edge-v3.0} & 1536 & 3072 & 0.7 & \makecell[tl]{\texttt{Wqkv, Wo, Wi,}\\\texttt{linear\_1, linear\_2,}\\\texttt{mlp.0, mlp.2, mlp.4}} \\
\addlinespace
\texttt{gliclass-modern-base-v3.0} & 512  & 1024 & 0.7 & \makecell[tl]{\texttt{Wqkv, Wo, Wi,}\\\texttt{linear\_1, linear\_2}} \\
\addlinespace
\texttt{gliclass-modern-large-v3.0} & 768  & 1536 & 0.7 & \makecell[tl]{\texttt{Wqkv, Wo, Wi,}\\\texttt{linear\_1, linear\_2}} \\
\addlinespace
\texttt{gliclass-base-v3.0} & 384  & 768  & 0.7 & \makecell[tl]{\texttt{query\_proj, key\_proj,}\\\texttt{value\_proj, dense,}\\\texttt{linear\_1, linear\_2,}\\\texttt{mlp.0, mlp.2, mlp.4}} \\
\addlinespace
\texttt{gliclass-large-v3.0} & 384  & 768  & 0.7 & \makecell[tl]{\texttt{query\_proj, key\_proj,}\\\texttt{value\_proj, dense,}\\\texttt{linear\_1, linear\_2,}\\\texttt{mlp.0, mlp.2, mlp.4}} \\
\bottomrule
\end{tabular}
\label{tab:lora-all-transposed}
\end{table}

\subsection{Evaluation}

We evaluate GLiClass models against strong cross-encoder baselines on standard text classification benchmarks including Rotten Tomatoes, CR, IMDB, and others (see Tables~\ref{tab:gliclass-v3} and~\ref{tab:cross-encoders} for complete results). We also report few-shot performance using 8 examples per label.

Inference speed is measured on a single NVIDIA A6000 GPU with batch size 1. We test across label counts $L \in \{1, 2, 4, 8, 16, 32, 64, 128\}$ and input lengths $T \in \{64, 256, 512\}$ tokens. For each $(L,T)$ configuration, we execute 10 forward passes and report average throughput in examples per second.

\section{Results}\label{seq:results}

Table~\ref{tab:gliclass-model-overview} summarizes model characteristics and performance. F1-score scales with model size within each family: \texttt{gliclass-large-v3.0} achieves the highest average (0.7193), followed by \texttt{base} (0.6764), \texttt{modern-large} (0.6197), \texttt{modern-base} (0.5577), and \texttt{edge} (0.4900). Throughput shows an inverse relationship: \texttt{edge} is fastest (97.29 ex/s), while \texttt{large} is slowest among GLiClass models (25.22 ex/s).

\begin{table}[ht]
\centering
\caption{GLiClass v3.0 Model Overview}
\label{tab:gliclass-model-overview}
\begin{tabular}{lcccc}
\toprule
Model name & Size & Params & Average Benchmark & Avg. Inference Speed (ex/s) \\
\midrule
gliclass-edge-v3.0        & 131 MB  & 32.7M  & 0.4900 & 97.29 \\
gliclass-modern-base-v3.0 & 606 MB  & 151M   & 0.5577 & 54.46 \\
gliclass-modern-large-v3.0& 1.6 GB  & 399M   & 0.6197 & 43.80 \\
gliclass-base-v3.0        & 746 MB  & 187M   & 0.6764 & 51.61 \\
gliclass-large-v3.0       & 1.75 GB & 439M   & 0.7193 & 25.22 \\
\bottomrule
\end{tabular}
\end{table}

Compared to cross-encoders (Table~\ref{tab:cross-encoders}), GLiClass achieves superior accuracy-latency trade-offs. \texttt{gliclass-large-v3.0} surpasses the strongest cross-encoder baseline (\texttt{deberta-v3-large-zeroshot-v2.0}, 0.6821) by +0.037 absolute (+5.5\% relative), while \texttt{gliclass-base-v3.0} remains within 0.006 absolute points. \texttt{gliclass-modern-large-v3.0} is comparable to \texttt{roberta-large-zeroshot-v2.0-c} (0.6197 vs. 0.6152).

\begin{table}[h]
\centering
\caption{Performance Comparison of GLiClass Models}
\label{tab:gliclass-v3}
\begin{tabular}{lcccccc}
\toprule
Dataset & \multicolumn{1}{p{1.8cm}}{\centering gliclass-large\\v3.0} & \multicolumn{1}{p{1.8cm}}{\centering gliclass-base\\v3.0} & \multicolumn{1}{p{1.8cm}}{\centering gliclass-modern-large\\v3.0} & \multicolumn{1}{p{1.8cm}}{\centering gliclass-modern-base\\v3.0} & \multicolumn{1}{p{1.8cm}}{\centering gliclass-edge\\v3.0} \\
\midrule
CR & 0.9398 & 0.9127 & 0.8952 & 0.8902 & 0.8215 \\
sst2 & 0.9192 & 0.8959 & 0.9330 & 0.8959 & 0.8199 \\
sst5 & 0.4606 & 0.3376 & 0.4619 & 0.2756 & 0.2823 \\
20\_news\_groups & 0.5958 & 0.4759 & 0.3905 & 0.3433 & 0.2217 \\
spam & 0.7584 & 0.6760 & 0.5813 & 0.6398 & 0.5623 \\
financial\_phrasebank & 0.9000 & 0.8971 & 0.5929 & 0.4200 & 0.5004 \\
imdb & 0.9366 & 0.9251 & 0.9402 & 0.9158 & 0.8485 \\
ag\_news & 0.7181 & 0.7279 & 0.7269 & 0.6663 & 0.6645 \\
emotion & 0.4506 & 0.4447 & 0.4517 & 0.4254 & 0.3851 \\
cap\_sotu & 0.4589 & 0.4614 & 0.4072 & 0.3625 & 0.2583 \\
rotten\_tomatoes & 0.8411 & 0.7943 & 0.7664 & 0.7070 & 0.7024 \\
massive & 0.5649 & 0.5040 & 0.3905 & 0.3442 & 0.2414 \\
banking & 0.5574 & 0.4698 & 0.3683 & 0.3561 & 0.0272 \\
snips & 0.9692 & 0.9474 & 0.7707 & 0.5663 & 0.5257 \\
\midrule
AVERAGE & \textbf{0.7193} & \textbf{0.6764} & \textbf{0.6197} & \textbf{0.5577} & \textbf{0.4900} \\
\bottomrule
\end{tabular}
\end{table}

\begin{table}[h]
\centering
\caption{Cross-Encoders Performance Comparison}
\label{tab:cross-encoders}
\begin{tabular}{lcccc}
\toprule
Dataset & \multicolumn{1}{p{2.2cm}}{\centering deberta-v3-large\\zeroshot-v2.0} & \multicolumn{1}{p{2.2cm}}{\centering deberta-v3-base\\zeroshot-v2.0} & \multicolumn{1}{p{2.2cm}}{\centering roberta-large\\zeroshot-v2.0-c} & \multicolumn{1}{p{2.2cm}}{\centering comprehend\_it\\base} \\
\midrule
CR & 0.9134 & 0.9051 & 0.9141 & 0.8936 \\
sst2 & 0.9272 & 0.9176 & 0.8573 & 0.9006 \\
sst5 & 0.3861 & 0.3848 & 0.4159 & 0.4140 \\
enron\_spam & 0.5970 & 0.4640 & 0.5040 & 0.3637 \\
financial\_phrasebank & 0.5820 & 0.6690 & 0.4550 & 0.4695 \\
imdb & 0.9180 & 0.8990 & 0.9040 & 0.4644 \\
ag\_news & 0.7710 & 0.7420 & 0.7450 & 0.6016 \\
emotion & 0.4840 & 0.4950 & 0.4860 & 0.4165 \\
cap\_sotu & 0.5020 & 0.4770 & 0.5230 & 0.3823 \\
rotten\_tomatoes & 0.8680 & 0.8600 & 0.8410 & 0.4728 \\
massive & 0.5180 & 0.5200 & 0.5200 & 0.3314 \\
banking77 & 0.5670 & 0.4460 & 0.2900 & 0.4972 \\
snips & 0.8340 & 0.7477 & 0.5430 & 0.7227 \\
\midrule
AVERAGE & \textbf{0.6821} & \textbf{0.6559} & \textbf{0.6152} & \textbf{0.5331} \\
\bottomrule
\end{tabular}
\end{table}

Few-shot adaptation with 8 examples per label consistently improves performance (Table~\ref{tab:gliclass-v3-few-shot}). Average gains over zero-shot are substantial: +0.1888 for \texttt{edge} (+50.0\%), +0.2094 for \texttt{modern-base} (+47.1\%), +0.1877 for \texttt{modern-large} (+36.1\%), +0.1067 for \texttt{base} (+18.4\%), and +0.1063 for \texttt{large} (+17.1\%). These results indicate that smaller variants benefit disproportionately from limited supervision.

\begin{table}[h]
\centering
\caption{GLiClass Model Performance in Zero-shot and Few-shot Learning}
\small
\label{tab:gliclass-v3-few-shot}
\begin{tabular}{p{3.5cm}cccccccccc}
\toprule
\textbf{Model} &
{\scriptsize\rotatebox[origin=c]{90}{\parbox{1.0cm}{\centering \textbf{Examples per label}}}} &
{\scriptsize\rotatebox[origin=c]{90}{\parbox{1.0cm}{\centering \textbf{sst5}}}} &
{\scriptsize\rotatebox[origin=c]{90}{\parbox{2.7cm}{\centering \textbf{financial\_phrasebank}}}} &
{\scriptsize\rotatebox[origin=c]{90}{\parbox{1.1cm}{\centering \textbf{ag\_news}}}} &
{\scriptsize\rotatebox[origin=c]{90}{\parbox{1.1cm}{\centering \textbf{emotion}}}} &
{\scriptsize\rotatebox[origin=c]{90}{\parbox{1.1cm}{\centering \textbf{cap\_sotu}}}} &
{\scriptsize\rotatebox[origin=c]{90}{\parbox{1.6cm}{\centering \textbf{rotten\_tomatoes}}}} &
{\scriptsize\rotatebox[origin=c]{90}{\parbox{1.1cm}{\centering \textbf{massive}}}} &
{\scriptsize\rotatebox[origin=c]{90}{\parbox{1.2cm}{\centering \textbf{banking77}}}} &
\textbf{Avg} \\
\midrule
gliclass-edge-v3.0 & 0 & 0.2779 & 0.4986 & 0.6669 & 0.3854 & 0.2306 & 0.6955 & 0.2389 & 0.0255 & 0.3774 \\
gliclass-edge-v3.0 & 8 & 0.3882 & 0.6998 & 0.7648 & 0.3989 & 0.3440 & 0.7344 & 0.5347 & 0.6644 & \textbf{0.5662} \\
\addlinespace
gliclass-modern-base-v3.0 & 0 & 0.2765 & 0.4199 & 0.6673 & 0.4237 & 0.3591 & 0.7070 & 0.3443 & 0.3581 & 0.4445 \\
gliclass-modern-base-v3.0 & 8 & 0.3947 & 0.8675 & 0.7742 & 0.4700 & 0.4363 & 0.8264 & 0.6937 & 0.7683 & \textbf{0.6539} \\
\addlinespace
gliclass-modern-large-v3.0 & 0 & 0.4629 & 0.5940 & 0.7268 & 0.4506 & 0.4115 & 0.7653 & 0.3876 & 0.3653 & 0.5205 \\
gliclass-modern-large-v3.0 & 8 & 0.5070 & 0.9066 & 0.8307 & 0.5337 & 0.4556 & 0.8638 & 0.7331 & 0.8354 & \textbf{0.7082} \\
\addlinespace
gliclass-base-v3.0 & 0 & 0.3377 & 0.8971 & 0.7279 & 0.4450 & 0.4681 & 0.7943 & 0.5041 & 0.4689 & 0.5804 \\
gliclass-base-v3.0 & 8 & 0.4324 & 0.9116 & 0.8295 & 0.4931 & 0.4867 & 0.8450 & 0.7008 & 0.7975 & \textbf{0.6871} \\
\addlinespace
gliclass-large-v3.0 & 0 & 0.4627 & 0.9000 & 0.7183 & 0.4501 & 0.4666 & 0.8411 & 0.5651 & 0.5575 & 0.6202 \\
gliclass-large-v3.0 & 8 & 0.5046 & 0.9042 & 0.8413 & 0.5303 & 0.5372 & 0.8827 & 0.7549 & 0.8563 & \textbf{0.7265} \\
\bottomrule
\end{tabular}
\end{table}

\begin{figure}[ht]
    \centering
    \includegraphics[width=1\linewidth]{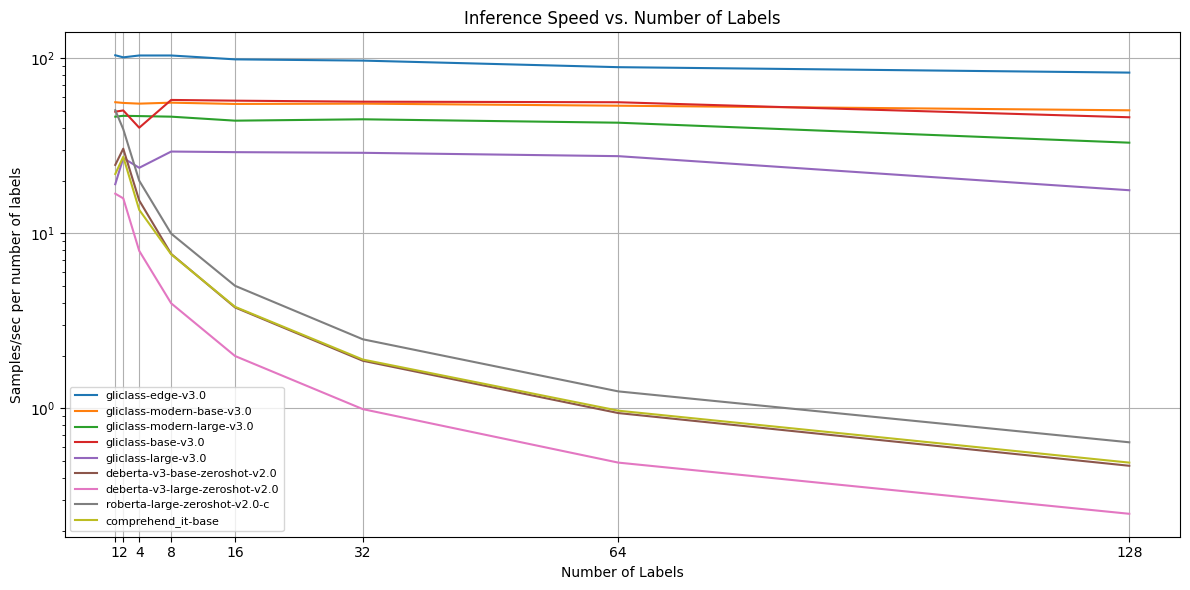}
    \caption{Models Inference Speed Comparison}
    \label{fig:inference-speed}
\end{figure}

GLiClass demonstrates superior scalability with increasing label counts (Table~\ref{tab:inference-speed}, Figure~\ref{fig:inference-speed}). For \texttt{gliclass-edge-v3.0}, throughput decreases modestly from 103.81 to 82.64 ex/s when scaling from 1 to 128 labels (-20\%). \texttt{gliclass-base-v3.0} drops by ~7\% (49.42→45.94 ex/s) and \texttt{gliclass-large-v3.0} by ~7.6\% (19.05→17.60 ex/s). In contrast, cross-encoders show dramatic degradation: \texttt{deberta-v3-base-zeroshot-v2.0} drops from 24.55 to 0.47 ex/s ($\approx 52\times$ slower). 

In aggregate, GLiClass delivers roughly $2.3\times$–$16\times$ higher average throughput than cross-encoders under our settings (e.g., \texttt{large} vs. \texttt{deberta-v3-base}: $25.22/10.63=2.37\times$; \texttt{edge} vs. \texttt{deberta-v3-large}: $97.29/6.03=16.1\times$; \texttt{base} vs. \texttt{roberta-large}: $51.61/16.12=3.2\times$).

Dataset-level variability is present (Table~\ref{tab:gliclass-v3}). \texttt{gliclass-large-v3.0} generally leads, while smaller or modern variants occasionally match or exceed it on specific tasks (e.g., \texttt{ag\_news} favors \texttt{base}; \texttt{sst5} is tight between \texttt{modern-large} and \texttt{large}). This suggests complementary inductive biases that can be further exploited in downstream selection.

\begin{table}[h]
\centering
\caption{Inference Speed: Samples per Second by Number of Labels (on A6000 GPU)}
\small
\label{tab:inference-speed}
\begin{tabular}{lccccccccc}
\toprule
\multicolumn{1}{p{3.5cm}}{Model Name} & 1 & 2 & 4 & 8 & 16 & 32 & 64 & 128 & \textbf{Average} \\
\midrule
gliclass-edge-v3.0 & 103.81 & 101.01 & 103.50 & 103.50 & 98.36 & 96.77 & 88.76 & 82.64 & \textbf{97.29} \\
gliclass-modern-base-v3.0 & 56.00 & 55.46 & 54.95 & 55.66 & 54.73 & 54.95 & 53.48 & 50.34 & \textbf{54.46} \\
gliclass-modern-large-v3.0 & 46.30 & 46.82 & 46.66 & 46.30 & 43.93 & 44.73 & 42.77 & 32.89 & \textbf{43.80} \\
gliclass-base-v3.0 & 49.42 & 50.25 & 40.05 & 57.69 & 57.14 & 56.39 & 55.97 & 45.94 & \textbf{51.61} \\
gliclass-large-v3.0 & 19.05 & 26.86 & 23.64 & 29.27 & 29.04 & 28.79 & 27.55 & 17.60 & \textbf{25.22} \\
deberta-v3-base-zeroshot-v2.0 & 24.55 & 30.40 & 15.38 & 7.62 & 3.77 & 1.87 & 0.94 & 0.47 & \textbf{10.63} \\
deberta-v3-large-zeroshot-v2.0 & 16.82 & 15.82 & 7.93 & 3.98 & 1.99 & 0.99 & 0.49 & 0.25 & \textbf{6.03} \\
roberta-large-zeroshot-v2.0-c & 50.42 & 39.27 & 19.95 & 9.95 & 5.01 & 2.48 & 1.25 & 0.64 & \textbf{16.12} \\
comprehend\_it-base & 21.79 & 27.32 & 13.60 & 7.58 & 3.80 & 1.90 & 0.97 & 0.49 & \textbf{9.72} \\
\bottomrule
\end{tabular}
\end{table}
\normalsize
\section{Discussion}\label{seq:discussion}
GLiClass effectively balances accuracy and speed, making it a versatile choice for sequence classification tasks. As model size grows from \texttt{edge} to \texttt{large}, the average F1-score rises significantly (from 0.4900 to 0.7193), while throughput decreases moderately (from 97.29 to 25.22 examples per second on an A6000 GPU). Unlike cross-encoders, which experience severe slowdowns with more labels (e.g., 50$\times$ slower from 1 to 128 labels), GLiClass maintains high efficiency, with only a slight throughput reduction (7--20\% from 1 to 128 labels). This efficiency comes from processing all labels in a single forward pass, ideal for production environments with large label sets. However, for very large label sets (e.g., 1000+), efficiency may drop due to context length limits (around 1024 tokens), potentially requiring techniques like truncation or batching.
Additionally, performance can degrade with larger label sets, as seen in datasets like \texttt{banking77}, where accuracy slightly declines. Our tailored training approach has enabled GLiClass to match cross-encoder performance despite these challenges, though cross-encoders handle dense information better. We attribute GLiClass's limitations with extremely large label sets to current positional encoding and attention mechanisms, which struggle to generalize across large contexts and effectively aggregate label information. These findings suggest opportunities for future research into improved positional encoding and attention mechanisms to enhance scalability for complex classification tasks.

The strong few-shot learning capabilities of GLiClass, particularly in smaller variants, underscore its adaptability to new domains. With just 8 examples per label, the \texttt{edge} and \texttt{modern-base} variants achieve substantial F1-score improvements (approximately 50\% relative gain), making them ideal for resource-constrained scenarios. This adaptability is driven by the joint text-label encoding strategy, which leverages contextual interactions to generalize from minimal supervision.

Table~\ref{tab:model-comparison} compares GLiClass with large language models (LLMs), cross-encoders, and embeddings-based models. GLiClass achieves better scalability and efficiency than cross-encoders. Still, further increasing the label sets can become more challenging for the model. We hypothesize it to the limitations of modern positional encoding and attention mechanisms. In the case of GLiClass, the task becomes more complex with the model; it should be generalized well to increase context size and aggregate information to label tokens. We believe that our work on GLiClass can inspire further work on better positional encoding and attention mechanism approaches. 
\begin{table}[ht]
\centering
\caption{Comparison of GLiClass, LLMs, Cross-encoders, and Embeddings Models for Classification Tasks}
\small
\label{tab:model-comparison}
\begin{tabular}{p{2.5cm}p{3.25cm}p{3.25cm}p{3.25cm}p{3.25cm}}
\toprule
\textbf{Aspect} & \textbf{GLiClass} & \textbf{LLMs} & \textbf{Cross-encoders} & \textbf{Embeddings Models} \\
\midrule
\textbf{Scaling with Number of Labels} & Non-linear; mild throughput decrease (e.g., $\sim$7--20\% from 1 to 128 labels) due to joint processing in single forward pass & Moderate; prompt length increases with labels, but generation time relatively constant unless very large sets & Poor; linear decrease in throughput as processes text-label pairs sequentially (e.g., 50$\times$ slower from 1 to 128 labels) & Excellent; constant time for text encoding, sub-linear for similarity computations (very fast even for large sets) \\
\addlinespace
\textbf{Performance Stability with Many Labels (e.g., 100+)} & 
Moderate; feasible up to context length limits (e.g., $\sim$1024 tokens), with efficiency maintained via a single pass, though truncation or batching may be required in extreme cases & 
Moderate; constrained by context window size (e.g., 8K--128K tokens), requires prompt engineering; efficiency decreases with very long prompts & 
Good accuracy due to pairwise computations, but inference time scales linearly with the number of labels & 
Excellent; maintains both high accuracy and computational efficiency \\
\addlinespace
\textbf{Computational Efficiency} & High; single pass for multi-label, comparable to embeddings, optimized for production (25--97 ex/s on A6000 GPU) & Low; autoregressive generation is computationally intensive, high latency for inference & Medium to Low; efficient per pair but scales poorly with labels due to repeated forward passes & High; fast encoding and vector operations, minimal compute per inference \\
\addlinespace
\textbf{Zero-Shot Capability} & Strong; designed for flexibility, outperforms cross-encoders on benchmarks (e.g., avg. F1 0.49--0.72) & Strong but inconsistent; versatile but struggles with instruction adherence & Strong; good for NLI-style classification but limited by lack of cross-label info & Moderate; effective for semantic matching but weaker on logical constraints \\
\addlinespace
\textbf{Few-Shot Capability} & Excellent; significant gains with minimal examples (e.g., +17--50\% F1 with 8 examples/label), especially for smaller variants & Strong; in-context learning allows adaptation, but requires careful prompting & Moderate; can fine-tune but not optimized for few-shot without additional training & Moderate to Strong; methods like SetFit enable efficient few-shot but may not capture complex patterns \\
\addlinespace
\textbf{Handling Complex Logical/Semantic Constraints} & Strong; joint text-label interactions capture relations, hierarchies, and dependencies; enhanced by logic/NLI post-training & Strong; capable of complex reasoning but may require large models & Moderate; pairwise processing misses inter-label dependencies, affecting complex scenarios & Weak; struggles with logical constraints, relies on semantic similarity \\
\addlinespace
\textbf{Overall Accuracy-Efficiency Trade-off} & Superior; balances high accuracy (surpasses cross-encoders by $\sim$5.5\%) with embedding-like efficiency and better scalability & Versatile but inefficient; high accuracy potential offset by latency and inconsistency & Good accuracy but poor scalability limits practical use for large label sets & Efficient with good baseline accuracy, but lower in complex tasks compared to others \\
\bottomrule
\end{tabular}
\end{table}

Post-training with Low-Rank Adaptation (LoRA) and specialized data streams (logic/NLI and pattern-focused) effectively mitigates initial limitations, such as attention degradation at extreme label-to-text ratios. The layer-wise attention re-weighting mechanism further enhances information flow, contributing to robust performance across diverse datasets. Notably, higher-rank LoRA adapters improve training stability for the \texttt{edge} variant, suggesting that smaller models benefit from over-parameterization during fine-tuning.

Consistent performance across datasets enables deployment-driven model selection: the \texttt{large} variant (0.7193 F1) suits quality-critical applications, the \texttt{base} variant (0.6764 F1) offers a balanced trade-off, and the \texttt{edge} variant (0.4900 F1, 97.29 ex/s) excels in high-throughput scenarios. Dataset-specific variability highlights complementary inductive biases among variants, which can be leveraged for task-specific optimization. Despite these strengths, challenges remain, including calibration variability across datasets and sensitivity to extreme label-to-text ratios. These can be addressed through targeted post-training, such as fine-tuning on diverse datasets or refining LoRA configurations. Our findings on GLiClass suggest that limitations in scaling to large label sets may be linked to current positional encoding and attention mechanisms, paving the way for future research into more robust approaches. Future work will also explore optimizing attention mechanisms for extreme conditions and extending GLiClass to multilingual and domain-specific settings to enhance its applicability.

\section{Conclusion}

We introduced GLiClass, a label-conditioned encoder transformer-based family for sequence classification that successfully bridges the gap between accuracy and efficiency. The architecture achieves state-of-the-art results on standard benchmarks while maintaining throughput that scales favorably with label count, which is a critical advantage over pairwise cross-encoders.

Key contributions include:
\begin{itemize}
    \item A novel uni-encoder architecture that jointly processes text and labels, enabling rich cross-label interactions;
    \item Superior accuracy-latency trade-offs, with the largest variant surpassing strong baselines by 5.5\% while maintaining practical inference speeds;
    \item Excellent few-shot learning capabilities, particularly for smaller models (up to 50\% improvement with 8 examples);
    \item Robust scaling behavior with label count, maintaining 80\% of single-label throughput even with 128 labels;
    \item Adaptation of proximal policy optimization to multi-label classification, which improves generalization and enables training on data with limited label annotations or training with human feedback.
\end{itemize}

The GLiClass family offers flexible deployment options: \texttt{large} (0.7193 F1) for quality-critical scenarios, \texttt{base} (0.6764 F1) for balanced deployments, \texttt{modern} variants for specific architectures, and \texttt{edge} (0.4900 F1, 97.29 ex/s) for maximum throughput. Throughput degrades only mildly with the number of labels, contrasting with the sharp slowdowns observed for pairwise cross-encoders.

Few-shot adaptation with 8 examples per label consistently improves performance, with the largest relative gains on smaller models, enabling practical adaptation under tight annotation and latency budgets. Post-training with LoRA and logic/pattern-focused streams stabilizes training and mitigates degradation under extreme label-text ratios.

Limitations include residual calibration differences across datasets, sensitivity under extreme label-text lengths, and variability on fine-grained taxonomies. Future work will focus on improving calibration across datasets and extending to multilingual settings and new domains.

\section{Availability}

Models are available through the GLiClass Python library at:
\url{https://github.com/Knowledgator/GLiClass}

Pre-trained models can be downloaded from the Hugging Face repository at:
\url{https://huggingface.co/collections/knowledgator/gliclass-v3-687a2d211b89659da1e3f34a}

\section{Acknowledgments}
\textit{We sincerely thank Urchade Zaratiana, the creator of GLiNER, whose work and encouragement greatly inspired the development of GLiClass.}

\bibliographystyle{plainnat}  
\bibliography{main}  
\end{document}